\documentclass{style/new_tlp}
\usepackage{mathptmx}

\usepackage{todonotes}
\usepackage{xspace}
\usepackage{url}
\usepackage{booktabs}
\usepackage{multirow}
\usepackage{marvosym}
\usepackage{amssymb}
\usepackage{subfigure}
\usepackage[ruled,vlined,linesnumbered]{algorithm2e}

\usepackage[acronyms]{glossaries}
\newacronym{ichd3}{\textit{ICHD-3}}{\textit{International Classification of Headache Disorders - $3^{rd}$ edition}}
\newacronym{ichd2}{\textit{ICHD-II}}{\textit{International Classification of Headache Disorders - $2^{nd}$ edition}}
\newacronym{asp}{\xspace{ASP}}{\textit{Answer Set Programming}}
\newacronym{dlp}{\xspace{DLP}}{{Disjunctive Logic Programming}}
\newacronym{ichd}{\textit{ICHD}}{\textit{International Classification of Headache Disorders}}

\newcommand\lar{\ensuremath{\,\leftarrow\,}}

\newcommand{\GP}[1]{\ensuremath{Gr(#1)}\xspace}
\newcommand{\HB}[1]{\ensuremath{B_{#1}}\xspace}
\newcommand{\HU}[1]{\ensuremath{U_{#1}}\xspace}

\def\naf{\ensuremath{\raise.17ex\hbox{\ensuremath{\scriptstyle\mathtt{\sim}}}}\xspace}

\usepackage[inline]{enumitem}

\usepackage{listings}
\lstdefinestyle{asp-style}{
	language=Prolog,
	frame=lines,
	keywordstyle=\linespread{1.1}\small\ttfamily,
	basicstyle=\linespread{1.1}\small\ttfamily,
	breaklines=true,
}
\title[A logic-based decision support system for the diagnosis of headache disorders] {A logic-based decision support system for the diagnosis of headache disorders according to the ICHD-3 international classification\thanks{This work has been partially supported by MISE under the project ``ALCMEONE'' (NUM:F/050502/01-02-03/X32, CUP:B28117000430008) -- Horizon 2020 PON 2014-2020.}}

 \author[R. Costabile, G. Catalano, B. Cuteri, M. C. Morelli, N. Leone and M. Manna]
         {ROBERTA COSTABILE\\
            Department of Mathematics and Computer Science, University of Calabria, Italy\\
            \email{r.costabile@mat.unical.it}
            \and GELSOMINA CATALANO\\
            DLVSystem Srl, Rende, Italy\\
            \email{catalano@dlvsystem.com}
            \and BERNARDO CUTERI\\
            Department of Mathematics and Computer Science, University of Calabria, Italy\\
            \email{cuteri@mat.unical.it}
            \and MARIA CONCETTA MORELLI\\
            Department of Mathematics and Computer Science, University of Calabria, Italy\\
            \email{maria.morelli@unical.it}
            \and NICOLA LEONE, MARCO MANNA \\
            Department of Mathematics and Computer Science, University of Calabria, Italy\\
            \email{\{leone,manna\}@mat.unical.it}}

\jdate{}
\pubyear{}
\pagerange{\pageref{firstpage}--\pageref{lastpage}}
\doi{}

\newcommand{\nop}[1]{}
\submitted{January 2003}
\revised{January 2003}
\accepted{January 2003}

\begin{document}
\maketitle
\begin{abstract}
Decision support systems play an important role in medical fields as they can augment clinicians to deal more efficiently and effectively with complex decision-making processes.
In the diagnosis of headache disorders, however, existing approaches and tools are still not optimal.
On the one hand, to support the diagnosis of this complex and vast spectrum of disorders, the International Headache Society released in 1988 the International Classification of Headache Disorders (ICHD), now in its 3rd edition: a 200 pages document classifying more than 300 different kinds of headaches, where each is identified via a collection of specific nontrivial diagnostic criteria.
On the other hand, the high number of headache disorders and their complex criteria make the medical history process inaccurate and not exhaustive both for clinicians and existing automatic tools.
To fill this gap, we present \textsc{head-asp}, a novel decision support system for the diagnosis of headache disorders. Through a REST Web Service, \textsc{head-asp} implements a dynamic questionnaire that complies with \acrshort{ichd3} by exploiting two logical modules to reach a complete diagnosis while trying to minimize the total number of questions being posed to patients.
Finally, \textsc{head-asp} is freely available on-line
and it is receiving very positive feedback from the group of neurologists that is testing it.\\
{\em Under consideration for publication in Theory and Practice of Logic Programming.}

  \end{abstract}


%\tableofcontents

% ``  ''

\section{Introduction}
\subsection{Context and state-of-the-art}

Decision support systems (DSS) have been conceived for providing the ``information and analysis necessary for the decisions that must be made''~\cite{DBLP:journals/tods/Donovan76}. After almost 50 year,
DSSs are still evolving, and they play an important role in various application domains. For example,
%definitely play an important role in the clinical domain as
they can augment clinicians to deal more efficiently and effectively with complex decision-making processes such as diagnostics, disease management, and drug control~\cite{CDDS_nature_2020}. Since the seventies, most has been done both in medicine and computer science to make DSS more and more robust and reliable. But in some specific fields, such as the diagnosis of headache disorders, existing approaches and tools are still not optimal.

\begin{figure}[t!]
	\centering
	\includegraphics[width=0.84\textwidth]{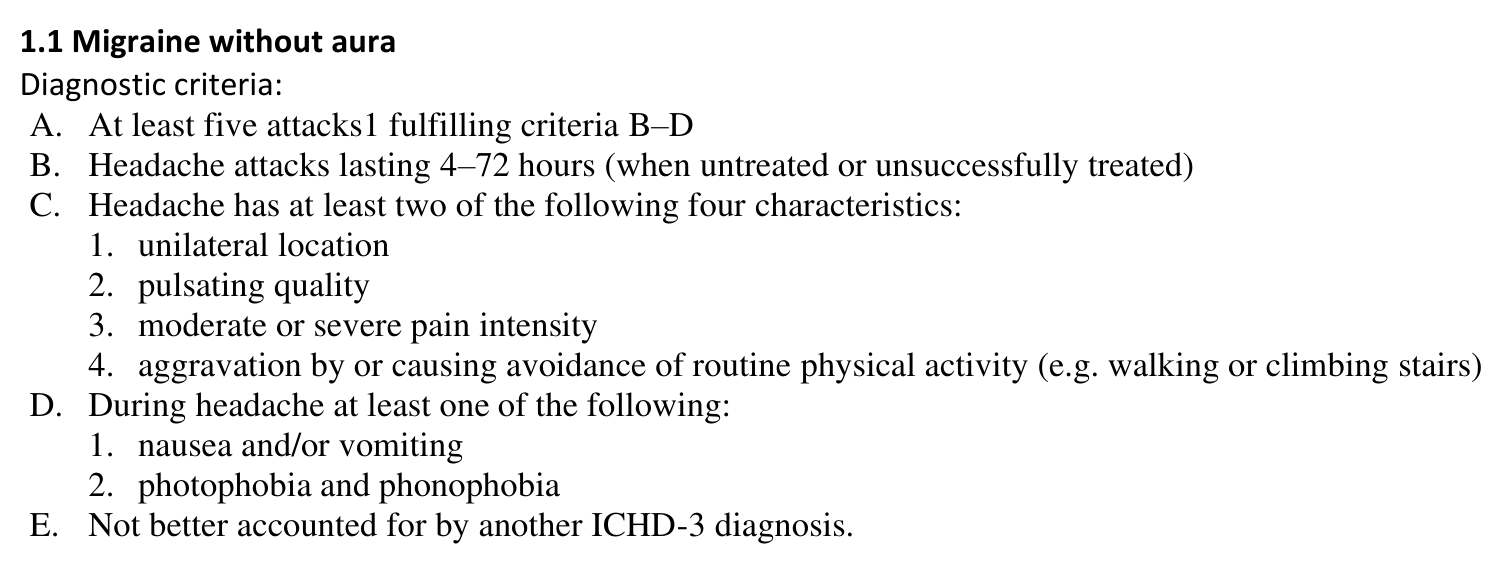}
	\caption{Diagnostic criteria of an ICHD-3 diagnosis.}
	\label{fig:migraine-without-aura}
\end{figure}

Headache disorders represent one of the most common and disabling conditions of the nervous system throughout the world~\cite{doi:10.1111/j.1468-2982.2007.01288.x}. In particular, about 90\% of all headaches are {\em primary}, namely, magnetic resonance imaging of the brain reveals no abnormality~\cite{doi:10.1111/head.13057}. To support the diagnosis of this complex and vast spectrum of disorders, in 1988, the {\em International Headache Society}~\citeyear{doi:10.1177/0333102417738202} released the first edition of the {\em International Classification of Headache Disorders} ({\em ICHD}), now in its 3rd edition: a document of 200 pages classifying, in a taxonomic way, more than 300 different kinds of headaches, and where each single form of headache is identified via a collection of specific nontrivial diagnostic criteria (see Figure~\ref{fig:migraine-without-aura}).
Typically, the diagnostic evaluation of headaches is mainly based on the description of symptoms by the patient. However, the medical history process may be inaccurate and not exhaustive, due to the high number of headache disorders identified by the medical community and characterized via the \acrshort{ichd3}. Thus, it is of paramount importance in this specific medical field to support clinicians and specialists during the entire diagnostic phase.

As said, a number of approaches in this domain have been already proposed in the literature. The most related ones are briefly discussed next.
\citeN{aida2} developed {\em AIDA Cefalee}, a system consisting of a database for the storage of symptoms and diagnostic data of patients paired with a module that can suggest possible diagnosis but only when all symptoms have been acquired. In particular, the database can be synchronized over the network allowing a continuous sharing of the patients' information and a cooperation between different research groups. The diagnostic tool has been validated experimentally but no details of the classification method are provided.
\citeN{simic2008computer} presented a novel tool that makes use of \emph{rule-based fuzzy logic} but is limited to a few forms of disorders. The researchers showed the workflow of the basic rule-based fuzzy logic systems model in which the rules are expressed as a collection of if-then statements. In particular, the information can be extracted by the patients in the form of if-then statements and these rules can be modeled using a fuzzy logic system; once the rules are provided to the system, it can be viewed as an input-to-output mapping.
\citeN{eslami2013computerized} proposed a DSS  implementing a dynamic questionnaire, but neither the system is publicly available nor the underlying classification method is described. In particular, the system provides questions related to headache disorders and, eventually, derives the most appropriate type of headache using simple human-like algorithmic logic; the accuracy of the diagnosis depends also on the accuracy of the patient's response.
\citeN{dong2014validation} proposed a general architecture of a DSS based on the \acrshort{ichd3} classification. In particular, the researchers'work is based on a 3-steps translation of the \acrshort{ichd3}: first in terms of flow-charts, then in terms of an ontological model and, finally, in terms of rules. The system is described mostly from an architectural point-of-view and in-depth details of the translation and of the diagnostic process are not provided.
%
%\citeN{yin2014clinical} designed a system based on a hybrid  intelligent reasoning approach that xxxxx; \todo{Roberta}
%
\citeN{vandewiele2018decision} proposed a DSS based on machine learning which generates an interpretable predictive model from the collected data. In particular, the system consists of three modules: a mobile application that captures symptomatic data from patients; an automated diagnosis support module that generates an interpretable decision tree, based on data semantically annotated with expert knowledge; and a web application that helps the clinicians to interpret captured data and learned insights by means of visualizations. The diagnostic process is based on supervised machine learning models.
%
%More details about these systems are given in Section~\ref{sec:related}.
%

\subsection{Motivation and objectives}

From the above overview, it should be already clear that none of the existing systems provides, at the same time,
$(i)$ the same level of accuracy required by \acrshort{ichd3},
$(ii)$ a solid, extensible and open knowledge representation model for fully and faithfully representing the \acrshort{ichd3} criteria,
$(iii)$ a dynamic questionnaire to support clinicians during the entire diagnostic phase, and
$(iv)$ an optimization strategy to minimize the number of questions posed to patients.
Moreover, $(v)$ none of the aforementioned systems is made available to be tested or used.

To make considerable steps forward in the diagnosis and management of headache disorders, the Italian Ministry of Economic Development appreciated and founded the research project {\em Alcmeone}, the main aim of which is providing an innovative organizational and management model, and an advanced technological platform of services for supporting the integrated clinical management of headache patients. In particular, concerning the headache diagnosis, the goal is to develop a decision support system that meets the following five main project specifications:
$(a)$ strictly represent \acrshort{ichd3} information, structure and criteria;
%$(b)$ represent also the implicit medical knowledge contained in the \acrshort{ichd3} guidelines;
$(b)$ focus on primary headaches, namely, on the first four chapters of the international classification;
$(c)$ implement an interactive questionnaire that rigorously guides both clinicians and patients during the medical history process;
$(d)$ reach a complete diagnostic picture of each patient by marking each primary headache diagnosis as compatible or not compatible;
$(e)$ keep reasonably low the number of questions posed to patients during the medical history process.

\subsection{Challenges and contribution}

Driven by the lack of effective tools in this domain and by the project specifications, we present in this paper \textsc{head-asp}, a novel decision support system for the diagnosis of headache disorders.
	During the development of the system, we faced three main technical challenges:
	designing a knowledge representation model being able to accommodate domain medical knowledge (often implicit in \acrshort{ichd3}) together with a natural and formal encoding of the \acrshort{ichd3} diagnoses (among the most complex in the medical field);
	designing a standard methodology to encode \acrshort{ichd3} diagnoses into logical rules over the data model mentioned above; and
	designing an efficient and effective heuristics offering a good trade-off between the average number of questions and the time required to determine the next question.

%\replace{Indeed,}{After more than two years of work,} \textsc{head-asp} fulfills \add{both the above five project specifications and} all the aforementioned desiderata:
%%
%$(i)$ it faithfully encodes all primary headache criteria specified in the \acrshort{ichd3} classification;
%%
%$(ii)$ the criteria of each diagnosis are expressed in a fully declarative way via \acrfull{asp} according to a formal logical model;
%%
%$(iii)$ the logical deductive module has been encapsulated in a REST Web service that implements a dynamic questionnaire to supports clinicians during diagnoses; and
%%
%$(iv)$ an additional ASP-based optimization module complements the deductive one in the Web Service to reduce the number of questions that are necessary to complete the diagnostic picture of patients.

After more than two years of work, \textsc{head-asp} does fulfill all the aforementioned desiderata and project specifications. From a technological viewpoint, the system consists of a REST Web service implementing a dynamic and interactive questionnaire that supports clinicians during the diagnostic phase. From a knowledge representation and reasoning perspective, the Web service encapsulates and manages two formal logical modules expressed in \acrfull{asp}: the first one is a deductive module that faithfully encodes all primary headache diagnoses and criteria of \acrshort{ichd3}, whereas the second one is an optimization module for minimizing the number of questions that are necessary to complete the diagnostic picture of patients.

Overall, we believe that \textsc{head-asp} fully meets the real needs in this domain.
This has been made possible thanks to the adoption of a declarative knowledge representation formalism and to a close interaction between clinicians and computer scientists.
Indeed, although it is still a research prototype, it is receiving very positive feedback from the group of neurologists that are testing and using it within {\em Alcmeone}. Some statistics on its effectiveness are reported in Section~\ref{sec:system}.
The system is freely available on-line at \url{https://head-asp.github.io/ichd-dss/}.

\section{Preliminaries on Answer Set Programming} %BERNARDO/MINA (bernardo aggiunge versione iniziale)
An  ASP \emph{atom} is of the form $p(t_1, \ldots, t_k)$, where $p$ is a predicate symbol and $t_1, \ldots, t_k$ are \emph{terms}. A term is either a constant or a variable. Variables start with an uppercase letter, and constants start with a lowercase letter.
A \emph{literal} is an atom $a_i$ (positive) or its negation $\mathtt{not}\ a_i$  (negative), where $\mathtt{not}$ denotes the \emph{negation as failure}.
An ASP program $\Pi$ is a finite set of rules of the form $head \lar body$, where $head$ is a disjunction of atoms $a_1 | \ldots | a_n$ with $n\geq 0$, and $body$ is a conjunction of literals $b_1,\ldots,b_m$ with $m\geq 0$.
A rule is called a \textit{fact} if it has an empty body and a \textit{constraint} if it has an empty head.
An object (atom, rule, etc.) is called {\em ground} if it contains no variables.
Rules and programs are \textit{positive} if they contain no negative literals, and \textit{general} otherwise.
Given $\Pi$, let the \emph{Herbrand Universe} \HU{\Pi} be the set of all constants appearing in $\Pi$ and the \emph{Herbrand Base} \HB{\Pi} be the set of all possible ground atoms which can be constructed from the predicate symbols appearing in $\Pi$ with the constants of \HU{\Pi}. Given a rule $r$, \GP{r} denotes the
set of rules obtained by applying all possible substitutions $\sigma$ from the variables in $r$ to elements of \HU{\Pi}. Similarly, given a program $\Pi$, the {\em ground instantiation} \GP{\Pi} of $\Pi$ is the set \( \bigcup_{r \in \Pi} \GP{r} \).

For every program, its answer sets are defined using its ground
instantiation in two steps: first answer sets of positive
programs are defined, then a reduction of general programs to positive
ones is given, which is used to define answer sets of general
programs.
An ASP interpretation $I$ for $\Pi$ is a subset of $\HB{\Pi}$.
An interpretation $M$ is a {\em model} for $\Pi$
if, for every $r \in \GP{\Pi}$, at
least one literal in the head of $r$ is true w.r.t.\ $M$ whenever all literals in the
body of $r$ are true w.r.t.\ $M$.
A model $X$ is an {\em answer set} (or {\em stable model})
for a positive program $\Pi$ if any other model $Y$ of $\Pi$ is such that ${X} \subseteq {Y}$.
The {\em reduct} or {\em Gelfond-Lifschitz transform}
of a general ground program $\Pi$ w.r.t.\ an interpretation $X$ is the positive
ground program $\Pi^X$, obtained from $\Pi$ by (i) deleting all rules
$r \in \Pi$ whose negative body is false w.r.t.\ X and (ii)
deleting the negative body from the remaining rules.
An answer set of %a general program
$\Pi$ is a model $X$ of $\Pi$ such
that $X$ is an answer set of $\GP{\Pi}^X$.

Over the years, the ASP language has been extended with additional constructs like strong negation, weak constraints, function symbols and aggregates~\cite{DBLP:journals/tplp/CalimeriFGIKKLM20}. In particular, in this work, we use strong negation and aggregate atoms. A strongly negated atom is an atom that starts with a strongly negation sign ($-$), in such case the atom is said to be {\em strongly negated}. The semantics of strongly negated atoms is the same of positive atoms (they can appear in heads and can be negated with $\mathtt{not}$ in bodies, etc.), with the addition that an {\em answer set} can not contain both an atom $a$ and the strongly negated atom $-a$.

Aggregates are also useful to model several rules in our context. More in details, an {\em aggregate atom} is an expression of the form $\#aggr\{t_1, \ldots, t_m : l_1, \ldots, l_n\} \odot  u$ where the expression $\#aggr \in \{\#count, \#sum, \#max, \#min\}$, $\odot \in \{<,  \leq, =, \neq, >, \geq\}$, $t_1, \ldots, t_m$ are terms, $l_1, \ldots, l_n$ are literals and $u$ is a variable or a number.
Intuitively, an aggregate is evaluated by first evaluating the aggregate body $l_1, \ldots, l_n$, which yields a set of ground tuples of $t_1, \ldots, t_m$, then the aggregation is computed on the multiset obtained by collecting all the first elements of all tuples and finally the result of the aggregation (according to the specific type of aggregation used) is compared with the term $u$, yielding the truth value of the aggregate.
%We refer to \cite{DBLP:journals/cacm/BrewkaET11} for a more comprehensive introduction to ASP.
% $f(S)$ is a function where $S$ is an expression of the form $\{\mathit{Vars}\! :\! \mathit{Conj}\}$,
%$\mathit{Vars}$ is a list of variables, $\mathit{Conj}$
%is a conjunction of atoms, and $f$ is an {\em aggregate function symbol}.
%The most common aggregate functions compute the number of terms, the sum of non-negative integers, and the minimum/maximum term in a set.
%As an example, the following rule counts the number of true instances of predicate $p$:
%$numP(X) \leftarrow \#count\{ X: p(X)\} = X$.

%\section{Overview of the System}%MARCO + BERNARDO

%\subsection{ICHD-3 International Classification}

%CAPPELLO AL PROBLEMA DA RISOLVERE  (OBIETTIVI DEL FRAMEWORK)
%DESCRIZIONE DELL'ICHD-3 CON SCREENSHOT E DISCUSSION
%MODALITà DI ANAMNESI CON QUESTIONARIO (VS ANAMESI STATICA)
%SPECIALIZZAZIONE RISPETTO AI PROFILI MEDICI

%\section{A Logic-Based Representation of ICHD-3}
\section{Knowledge base of the system}%ROBERTA + MINA
%\acrfull{ichd3} is considered by the World Health Organization as the official classification of headaches.
In this section we analyze the guidelines in detail identifying the essential aspects for their logic encoding and, then, we propose a representation in \acrshort{asp}.

\subsection{ICHD-3: basics}

\acrfull{ichd3} provides specific criteria, defined in natural language, for diagnosing known headaches. The possible diagnoses are organized in a hierarchical structure that expresses the existing relations between them. Each type of headache diagnosis includes its own sub-categories, that is, other more specific types of headache, which correspond to a higher level of detail.

The classification consists of 14 chapters grouped into 4 parts. This work focuses on primary headaches (part 1, chapters 1--4)
according to the project specifications reported above.  Anyway, the designed methodology can be definitely also applied to encode the rest of the diagnoses since they do not differ substantially in the structure.
%, which are classified in Part 1.
%In these cases, the pain occurs without documented causes, that is, headaches are not the result of another medical condition.
%Each chapter concerning primary headaches collects diagnoses related to a particular type of headache: \emph{Migraine}, \emph{Tension-type headache (TTH)}, \emph{Trigeminal autonomic cephalalgias (TACs)} and \emph{Other primary headache disorders}.
Primary headaches are subdivided into: ``\textsf{Migraine}'', ``\textsf{Tension-type headache (TTH)}'', ``\textsf{Trigeminal autonomic cephalalgias (TACs)}'', and ``\textsf{Other primary headache disorders}''.
In the following, we describe the structural aspects of the diagnoses, and then we report the main notions that underlie their content.

%\paragraph{Structural aspects.}
The \textit{diagnosis} represents the fundamental structural unit of \acrshort{ichd3}. Each diagnosis is identified by a set of \textit{criteria} that appear within a list marked with letters (``\textsf{A}'', ``\textsf{B}'', ...). Each criterion includes a series of requirements framed within the symptomatic state of the diagnosis the criterion refers to. A diagnosis is considered \emph{compatible} if, considering the ailments the patient suffers from, the conditions expressed by all its criteria are met.
A criterion can be presented in a monothetic or polythetic form.
A criterion is considered monothetic when it identifies a set of specific requirements. The necessary condition for it to be validated is that all its requirements are met. A criterion is considered polythetic when it consists of requirements that appear in an enumerative list format within its own statement.
To make such type of criterion as simple as possible, we assign the meaning of \textit{sub-criterion} to each set of requirements identified by an element of this numbered list (thus, each sub-criterion is marked with a number). The necessary condition for the validation of a polythetic criterion is to verify a minimum number of sub-criteria on the basis of a fixed inclusion threshold (see Figure~\ref{fig:migraine-without-aura}). A systematic analytical phase was necessary to identify the main notions of the classification. At the basis of \acrshort{ichd3} there is the notion of \textit{symptom}; it can be identified as a key notion because it is involved in the criteria of all the diagnoses of primary headaches.
Throughout the systematic analysis of the \acrshort{ichd3} contents, we extracted the \emph{attributes} associated with the symptoms:
%Essential information concerns:
%\begin{itemize}
    $(i)$ \emph{location of pain} (unilateral, bilateral, etc.);
    $(ii)$  \emph{aggravating factors} that worsen the pain (such as the movement, etc.) and any \emph{limitations} caused by pain (such as perform routine physical activities); and
    $(iii)$ \emph{type of pain} associated with headache (pulsating, intense, etc.).
%\end{itemize}
Furthermore, symptoms can be also characterized by:
    $(i)$ \emph{duration}, meant as the persistence over time of the pain caused by a symptom;
    $(ii)$ \emph{frequency} of pain attacks, i.e., the number of times the pain caused by a symptom occurs over a specified length of time (such as how many times a day, how many days a month the pain occurs) and, moreover, the (continuous) time interval in which a certain frequency lasts (as an example \textquotedblleft headache occurs on 1-14 days/month on average for \textgreater 3 months\textquotedblright);
    $(iii)$ \emph{number of attacks} that affect the patient; and
    $(iv)$ information relating to the report of a particular \emph{clinical exam} previously done by the patient.

%aggiungere malessere
%Finally, the most recurrent linguistic forms that require specific encoding methods have been identified in the diagnostic criteria  (for example,\textquotedblleft Headache has at least k of the following conditions ... \textquotedblright with k equal to one or two or other values).

%\begin{itemize}
 % \item  Descrizione attributi associati ai sintomi:
 % \item  Descrizione:numero attacchi, (durata, frequenza (frequenzaGiorniAlMese, frequenzaInPeriodoDiAttacchi, durataFrequenza), esame clinico
%\end{itemize}

%ELENCARE GLI ELEMENTI ESSENZIALI CHE CI SERVONO PER LA CODIFICA

\subsection{ICHD-3: logical representation}\label{sec:kr}
%DEFINIRE LO SCHEMA LOGICO (NOME PREDICATO, ARIETà, TERMINI CON SIGNIFICATO)
%Structural aspects
%Based on the structural analysis of \acrshort{ichd3} guidelines and showed in the previous section, we defined a formal representation in \acrshort{asp} that captures the knowledge they contain. The idea is that, given the disorders complained by the patient, by means of logical rules, it is possible to automatically identify the plausible diagnoses.

%\paragraph{Structural aspects encoding.}
In what follows, we encode \acrshort{ichd3} diagnoses and criteria via an ASP program $\mathcal{P}$ simply using (stratified) negation, aggregates and strong negation. This program builds on a core relational schema consisting of 18 predicates of three different types. Three of these predicates are intensional (type 3) and are used to derive diagnoses, criteria and subcriteria of a specific patient; the remaining ones are extensional. Among the latter, six of them (type 1) are used to represent key notions of \acrshort{ichd3}, such as all possible diagnoses, symptoms and their attributes; the remaining ones (type 2) are used to encode patients history, such as specific symptoms and the number of their attacks.
Hence, all instances of predicates of type 1 and type 3, as well as all rules, are always present in $\mathcal{P}$. Conversely, instances of predicates of type 2 vary with patients.
%
%We now present the all these predicates.

\paragraph{Type 1.}
As said, this group of predicates models key notions of \acrshort{ichd3}.
We first represent all possible diagnoses, symptoms and attributes via the following binary predicates:
\emph{ichdDiagnosis(Id, Name)},
\emph{ichdSymptom(Id, Name)} and \emph{ichdAttribute(Id, Name)}, where, in all the cases, the first term is an identifier and the second one its name. For example, \emph{ichdDiagnosis(d.1.1, \textquotedblleft migraine without aura'')} represents the diagnosis  ``\textsf{1.1. Migraine without aura}'', \emph{ichdSymptom(s4, \textquotedblleft headache'')} represents the symptom  ``\textsf{headache}'' identified by code\emph{ s4}, and \emph{ichdAttribute(loc2, \textquotedblleft bilateral location'')} represents the attribute ``\textsf{bilateral location}'' with its identification code \emph{loc2}.
Moreover, we also make explicit some properties of attributes that are only implicit in \acrshort{ichd3}. First, there are cases in which the presence of an attribute associated to a specific symptom excludes the possibility to have some other attribute for the same symptom. We use the predicate
\emph{mutuallyExclusive(Id\_attr\_1, Id\_attr\_2)} to model such kind of information. For example, if a headache is characterized by a unilateral location, then it cannot be characterized, at the same time, by a bilateral location. Hence, we may have \emph{mutuallyExclusive(loc1, loc2)} to express the mutual exclusion between the attribute ``\textsf{unilateral location}'' identified by code \emph{loc1} and the attribute ``\textsf{bilateral location}'' identified by code \emph{loc2}. Clearly, these two identifiers occur in two facts of the form: \emph{ichdAttribute(loc1, \textquotedblleft unilateral location'')} and \emph{ichdAttribute(loc2, \textquotedblleft bilateral location'')}.
Similarly, we also have \emph{sameAs(Id\_attr\_1, Id\_attr\_2)} to specify that, if an attribute is associated with a specific symptom, then, at the same time, also the attribute semantically equivalent to the first one should be associated with the same symptom. This is very important because \acrshort{ichd3} often uses different synonymous terms in different diagnostic criteria; without modeling such similarities our encoding would not completely reflect the intended \acrshort{ichd3} meaning.
For example, we include in $\mathcal{P}$ \emph{sameAs(int2, int10)}, \emph{ichdAttribute(int2, \textquotedblleft strong intensity'')}, and \emph{ichdAttribute(int10, \textquotedblleft acute intensity'')}.
Finally, we represent the dependence between diagnoses, symptoms and attributes via the predicate
\emph{isA(Id\_1, Id\_2)}. An example concerning the symptoms is the atom \emph{isA(s18, s54)}, where \emph{s18} and \emph{s54} are provided by \emph{ichdSymptom(s18, \textquotedblleft diplopia'')} and \emph{ichdSymptom(s54, \textquotedblleft visual symptom'')}.
%
%(i.e., the actual ICHD-3 taxonomy). This has been extensively done vi the predicate \emph{isA(Id\_sub\_diag, Name\_sub\_diag, Id\_diag, Name\_diag)}, where ..... For example,...

%-- later exploited to further optimize the dynamic questionnaire

\paragraph{Type 2.}
In this group we have predicates used to encode patients history. We start with \emph{symptom(Id\_sym)} and \emph{symAttribute(Id\_sym, Id\_attr)}, modelling the fact that a patient has a specific symptom and the fact that his/her symptoms have some peculiarity modeled via what we called attributes (e.g., symptom location, pain type); for example, \emph{symptom(s4)} and
\emph{symAttribute(s4, loc2)} represent that the patient reported the symptom ``\textsf{headache}'' with the attribute ``\textsf{bilateral location}''; indeed, \emph{ichdSymptom(s4, \textquotedblleft headache'')} and \emph{ichdAttribute(loc2, \textquotedblleft bilateral location'')} hold.
We also use the predicates
\emph{minAttacks(Id\_sym, Value)} and \emph{maxAttacks(Id\_sym, Value)} to specify that the actual number of attacks of a  patient's symptom falls in a certain range; for example {\emph{minAttacks(s4, 5)} (resp. \emph{maxAttacks(s4, 10)})}  indicates that the patient reported at least {``\textsf {5}''(resp. at most ``\textsf {10}'')}  attacks associated with the symptom identified by code \emph{s4}.
Similarly, we use the predicates \emph{minDuration(Id\_sym, Value)} and \emph{maxDuration(Id\_sym, Value)} to specify the duration of the pain caused by a certain symptom. As an example, \emph{minDuration(s4, 240)} (resp. \emph{maxDuration(s4, 4320)}) specify that the patient reported that the symptom \emph{s4} lasts at least \emph{240} (resp. at most \emph{4320}) minutes.
In the same way, frequency of pain attacks that are associated to a certain symptom is modeled by means of the predicates \emph{minDaysPerMonth(Id\_sym, Value)} and \emph{maxDaysPerMonth(Id\_sym, Value)}.
Finally, we use the predicate \emph{reportedCriterion(Description)}
to model some portions of text for which it was not possible and convenient to identify a decomposed modeling. Indeed, they express peculiar assertions only of certain diagnoses, occurring infrequently within the domain or, when reported, do not present syntactic or semantic variations. The only term of this predicate is the textual description of the statement. For example, \emph{reportedCriterion (\textquotedblleft At least one first or second degree family member has had attacks that meet the criteria of hemiplegic migraine'')} expresses the presence of the specified statement.
\paragraph{Type 3.}
Predicates in this group have the following signature: \emph{diagnosis(Id)}, \emph{criterion(Id\_diag, Letter)},
\emph{subCriterion(Id\_diag, Letter\_crit, Number)}.
They model the identifiers of diagnoses, criteria and subcriteria that can be derived for a specific patient.
As an example, if the atom \emph{diagnosis(d.1.1)} can be derived from $\mathcal{P}$, this means that the \acrshort{ichd3} diagnosis  ``\textsf{1.1. Migraine without aura}'' is compatible with patient history. Likewise, the derivation of the atom \emph{criterion(d.1.1,\textquotedblleft A'')}
means that criterion ``\textsf{A}'' of ``\textsf{1.1. Migraine without aura}'' is compatible with patient history, namely, the patient reported ``\textsf{at least five attacks fulfilling criteria B-D}''. Finally, the derivation of the atom \emph{subCriterion(d.1.1,\textquotedblleft C'',1)} means that subcriterion ``\textsf{1}'' of criterion ``\textsf{C}'' of ``\textsf{1.1. Migraine without aura}'' is compatible with patient history, namely, the patient reported that headache has the characteristic ``\textsf{unilateral location}''.

%
%the information as it is represented in \acrshort{ichd3}: the former contains the identification code and the denomination of the diagnosis, whereas the latter contains the identification code of the diagnosis it refers to, its marking letter (or label) and its informal description.
%As an example, the fact  \emph{diagnosis(\textquotedblleft d.1.1\textquotedblright,\textquotedblleft migraine without aura\textquotedblright)} refers to the diagnosis 1.1 called \textsf{Migraine without aura}. Likewise, the fact \emph{criterion(\textquotedblleft d.1.1\textquotedblright,\textquotedblleft A\textquotedblright, \textquotedblleft At least five attacks fulfilling criteria B-D\textquotedblright)}
%refers to criterion A related to the diagnosis 1.1, which requires at least five attacks fulfilling criteria B-D. \\

\begin{figure}[t!]
    \centering
    \includegraphics[width=0.7\textwidth,keepaspectratio]{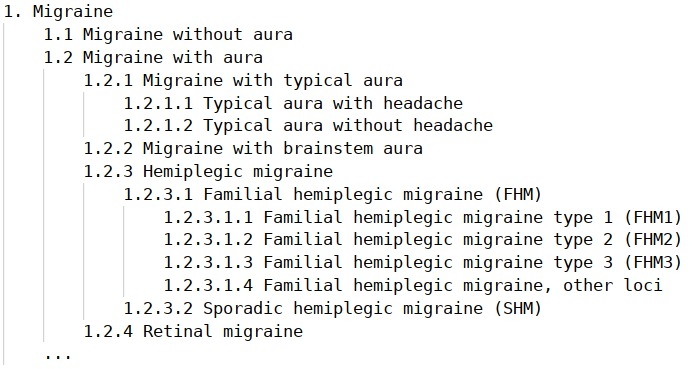}
    \caption{Some ICHD-3 diagnoses.}\label{fig:ichd3Tree} %\vspace{-0.3cm}
\end{figure}
\paragraph{Rules.}

%\noindent Preambolo e spiegazione primo gruppo di  regole
As shown in the previous subsection, the diagnoses are organized in a hierarchical structure  which expresses the specialization-generalization relations existing between them  (Figure~\ref{fig:ichd3Tree}).
 Intuitively, if a more specific diagnosis is compatible, then it is possible to infer a more generic diagnosis (the former being a sub-type of the latter); conversely, if a higher level diagnosis is not compatible, all its related specializations are invalidated. The following rules, among the others, express the implications inferable from the relations between diagnoses.
\begin{equation*}
	\begin{array}{l}
		r_1: diagnosis(Id\_sup) \leftarrow  diagnosis(Id),\  isA(Id,Id\_sup).\\
		r_2: -diagnosis(Id) \leftarrow -diagnosis(Id\_sup), isA(Id, Id\_sup).\\
	\end{array}
\end{equation*}
%\noindent Preambolo e spiegazione secondo gruppo di  regole
The two rules shown below define the conditions that must be met so that the diagnosis ``\textsf{1.1. Migraine without aura}'' can be compatible or not compatible (Figure~\ref{fig:migraine-without-aura}).
Intuitively, rule $r_3$ derives the diagnosis as true (compatible) if all its criteria ``\textsf{(A-D)}'' are true.
Similarly, rule $r_4$ expresses that the diagnosis is certainly false (not compatible) if at least one of its criteria is certainly false. Note that to use the strong negation is quite convenient to model such scenarios in which we need to distinguish criteria and diagnoses that are definitely confirmed from those definitely disconfirmed and from those still undefined, without any need in adding new predicates and further constants. This is also reflected to the predicates encoding patient's symptomatology.
\begin{equation*}
	\begin{array}{l}
		r_3:   diagnosis(Id) \leftarrow ichdDiagnosis(Id, \emph{\textquotedblleft migraine without aura\textquotedblright}), \\
		\ \ \ \ \ \ \ \ \ \ \ \
		criterion(\emph{Id, \textquotedblleft A''}),\ criterion(\emph{Id, \textquotedblleft B''}), \
		criterion(\emph{Id, \textquotedblleft C''),} \ criterion(\emph{Id, \textquotedblleft D'')}. \\
		r_4:  -diagnosis(Id)\leftarrow ichdDiagnosis(Id, \emph{\textquotedblleft migraine without aura\textquotedblright}),  -criterion(Id, \_).
	\end{array}
\end{equation*}
%\noindent Preambolo e spiegazione terzo gruppo di  regole
Each criterion (resp. subcriterion) of \acrshort{ichd3} is encoded through rules whose head is constituted by the predicate \emph{criterion} (resp. \emph{subcriterion}), while the body models effectively the semantics of the statement of the criterion (resp. sub-criterion), on the basis of its informal description in the international classification. As an example, consider criterion ``\textsf{B}" of diagnosis ``\textsf{1.1}'': ``\textsf{Headache attacks lasting 4-72 hr}''.
The following rules model such monothetic criterion. Declaratively, rule $r_5$ states that if the symptom ``\textsf{headache}''  is true and its duration respects the specified time interval, then \emph{criterion(d.1.1,\textquotedblleft B'')} will be derived as true. Rules $r_6$, $r_7$, $r_8$ express the conditions according to which the criterion is certainly false.
\begin{equation*}
	\begin{array}{l}
		
		r_5: criterion(Id,\emph{\textquotedblleft B''}) \leftarrow ichdDiagnosis(Id, \emph{\textquotedblleft migraine without aura\textquotedblright}),~~~~~~~~~~~~~~~~~~~~~~~~~~~~~~~~~~~~~~~ \\

		\ \ \ \ \ \ \ \ \ \ \ \
		ichdSymptom(Id\_sym,\emph{\textquotedblleft headache\textquotedblright}), \ symptom(Id\_sym), \ minDuration (Id\_sym,240), \\
		\ \ \ \ \ \ \ \ \ \ \ \
		 maxDuration(Id\_sym,4320).\\

r_6: -criterion(Id,\emph{\textquotedblleft B''}) \leftarrow ichdDiagnosis(Id, \emph{\textquotedblleft migraine without aura\textquotedblright}), \\
			\ \ \ \ \ \ \ \ \ \ \ \ ichdSymptom(Id\_sym,\emph{\textquotedblleft headache\textquotedblright}),\ symptom(Id\_sym), \  -minDuration (Id\_sym,240). 
	\end{array}
\end{equation*}
\begin{equation*}
	\begin{array}{l}

		r_7: -criterion(Id,\emph{\textquotedblleft B''}) \leftarrow ichdDiagnosis(Id, \emph{\textquotedblleft migraine without aura\textquotedblright}), \\
		\ \ \ \ \ \ \ \ \ \ \ \ ichdSymptom(Id\_sym,\emph{\textquotedblleft headache\textquotedblright}),\ symptom(Id\_sym), \  -maxDuration (Id\_sym,4320). \\
		
		r_8: -criterion(Id,\emph{\textquotedblleft B''}) \leftarrow ichdDiagnosis(Id, \emph{\textquotedblleft migraine without aura\textquotedblright}), \\
		\ \ \ \ \ \ \ \ \ \ \ \ ichdSymptom(Id\_sym,\emph{\textquotedblleft headache\textquotedblright}),\ -symptom(Id\_sym).

	\end{array}
\end{equation*}
%\noindent Preambolo e spiegazione quarto gruppo di  regole
Consider criterion ``\textsf{C}" of diagnosis ``\textsf{1.1}'': ``\textsf{Headache has at least two of the following four characteristics: $(i)$ unilateral location; $(ii)$ pulsating quality; $(iii)$ moderate or severe pain intensity; $(iv)$ aggravation by or causing avoidance of routine physical activity (eg, walking or climbing stairs)}''.
The following portion of the program corresponds to the encoding of such polythetic criterion.
\begin{equation*}
	\begin{array}{l}

     r_{9}: criterion(Id,\emph{\textquotedblleft C''}) \leftarrow ichdDiagnosis(Id, \emph{\textquotedblleft migraine without aura\textquotedblright}), \\
             \ \ \ \ \ \ \ \ \ \ \ \  \#count\{X:  subCriterion(Id, \emph{\textquotedblleft C'', X)\} }>= 2.\\
    r_{10}: -criterion(Id,\emph{\textquotedblleft C''}) \leftarrow ichdDiagnosis(Id, \emph{\textquotedblleft migraine without aura\textquotedblright}), \\
             \ \ \ \ \ \ \ \ \ \ \ \  \#count\{X:  -subCriterion(Id, \emph{\textquotedblleft C'', X)\} }>= 3.
\\

    r_{11}: subCriterion(Id,\emph{\textquotedblleft C''},1)  \leftarrow ichdDiagnosis(Id, \emph{\textquotedblleft migraine without aura\textquotedblright}), \\
    \ \ \ \ \ \ \ \ \ \ \ \
    ichdSymptom(Id\_sym,\emph{\textquotedblleft headache\textquotedblright}), \	symptom(Id\_sym), \\
	\ \ \ \ \ \ \ \ \ \ \ \
    symAttribute(Id\_sym, Id\_attr), \ ichdAttribute(Id\_attr, \emph{\textquotedblleft unilateral location\textquotedblright}).\\

    r_{12}: -subCriterion(Id,\emph{\textquotedblleft C''},1)  \leftarrow ichdDiagnosis(Id, \emph{\textquotedblleft migraine without aura\textquotedblright}), \\
		\ \ \ \ \ \ \ \ \ \ \ \ ichdSymptom(Id\_sym,\emph{\textquotedblleft headache\textquotedblright}),\ -symptom(Id\_sym). \\

  r_{13}: -subCriterion(Id,\emph{\textquotedblleft C''},1)  \leftarrow ichdDiagnosis(Id, \emph{\textquotedblleft migraine without aura\textquotedblright}), \\
    \ \ \ \ \ \ \ \ \ \ \ \
    ichdSymptom(Id\_sym,\emph{\textquotedblleft headache\textquotedblright}), \	symptom(Id\_sym), \\
	\ \ \ \ \ \ \ \ \ \ \ \
    -symAttribute(Id\_sym, Id\_attr), \ ichdAttribute(Id\_attr, \emph{\textquotedblleft unilateral location\textquotedblright}).\\

\end{array}
\end{equation*}
In this case, for a more natural modeling, it was convenient to use the aggregation construct \emph{\#count}. In particular, the rule $r_9$ aggregates the satisfied subcriteria and counts them (\emph{\#count}) in order to check if the resulting value matches the number required by the criterion itself.
It can be noted that the rules that express the conditions according to which the predicate in the head is certainly false can be automatically generated by denying, one at a time, the predicates of type 2.

\section{The next question strategy}%ROBERTA + MARIA CONCETTA
%STRATEGIA  + PROGRAMMA DISJUNTIVO
%can we directly say ``the questionnaire''?
As said in the Introduction, the goal is to develop a decision support system implementing a questionnaire that: $(i)$ rigorously guides both clinicians and patients during the medical history process; $(ii)$ automatically adapts to patients; and $(iii)$ reaches, in a reasonable amount of time, a complete diagnostic picture by inferring every diagnosis as either {\em compatible} or {\em not compatible}.
To this end, to satisfy $(i)$, we formulate each possible question as binary (i.e., their answers are simply ``yes'' or ``no'') to overcome the typical fact that patients are not always able to adequately describe their disorders.
Moreover, we design a logical module ---on top of the \acrshort{ichd3} encoding of Section~\ref{sec:kr}--- that, at each step of the questionnaire, identifies the convenient {\em next question} described in the following.
To satisfy $(ii)$, we discard from the set of candidate next questions those that are inappropriate or irrelevant (an example of an inappropriate question is a question related to the presence of the symptom ``\textsf{diplopia}'' if the patient already reported not to have the symptom ``\textsf{visual disorder}''; an example of an irrelevant question is a question concerning a diagnosis that has already been inferred as \emph{not compatible}.)
To satisfy $(iii)$, we implement a greedy strategy that first computes, for each candidate next question, the minimum number of inferred diagnoses  when considering both the ``yes'' or ``no'' answer and then, among these values, it selects the maximum one along with its associated question.

\paragraph{Architecture of the questionnaire.}

%%%%%%%%%%%%%%%%%%%%%%%%%% DIAGNOSTIC PROCESS
\begin{algorithm}[t]
	\small
	\SetKwInOut{Output}{Output}
	\SetKwInOut{Input}{Input}
	\Input{An ASP encoding $\Pi = \Pi_{ichd} \cup \Pi_q$, where $\Pi_{ichd}$ is the encoding that models the ICHD guidelines and $\Pi_q$ is the encoding that implements the questionnaire logic}
	\Output{A set of {\em compatible} diagnosis $C$, a set of {\em not compatible} diagnosis $N$, s.t. $C \cup N$ is the set of all diagnosis encoded in $\Pi_{ichd}$}
	\Begin{
		$U = possibleDiagnosis(\Pi_{ichd})$\\ \label{ln:init-U}
		$H =  \emptyset$, $C = N = \emptyset$\\ \label{ln:init-C-I}
		\While{$C \cup N \neq U$}{ \label{ln:loop}
			$\Pi^\prime = \Pi_{ichd} \cup H$\\
			$S = solveUnique(\Pi^\prime)$\\ \label{ln:solve-unique}
			$C = \{c \in S \mid predicate(c) = diagnosis \land isPositive(c)\}$\\
			$N = \{c \in S \mid predicate(c) = diagnosis \land isNegative(c)\}$\\
			\If{$C \cup N \neq U$}{ \label{ln:unfinished}
				$\Pi'' = \Pi_{ichd} \cup \Pi_q \cup H$\\
				$\mathcal{S} = solve(\Pi'')$\\ \label{ln:solve}
				$Q = \bigcup_{S \in \mathcal{S}}{\{c \in S \mid predicate(c) = ask\}}$\\ \label{ln:q}
				$B_{score} = -1$, $B_q = \bot$\\ \label{ln:init-score}
				\ForAll{$q \in Q$}{								
					$d_{yes} = getDeterminedDiagnosisCount(q, yes, \mathcal{S})$\\ \label{ln:answer-yes}
					$d_{no} = getDeterminedDiagnosisCount(q, no, \mathcal{S})$\\ \label{ln:answer-no}
					$q_{score} = min(d_{yes}, d_{no})$\\
					\If{$q_{score}  > B_{score}$}{ \label{ln:best-start}
					 	$B_{score} =q_{score}$, $B_q = q$\\ \label{ln:maxmin-end}
					}						
				}
				$H = H \cup patientAnswer(B_q)$\\	 \label{ln:add-patient-answer}			
			}
		}
		\Return{C, N}
	}
	\caption{Questionnaire process}\label{alg:questionnaire}
\end{algorithm}
%%%%%%%%%%%%%%%%%%%%%%%%%%
%Algorithm [?] represents the questionnaire in pseudo-code.
%BERNARDO algorihm/figure+formalization

Algorithm \ref{alg:questionnaire} presents the questionnaire process as implemented in our system.
The algorithm takes as input the ASP encoding of \acrshort{ichd3} ($\Pi_{ichd}$), as described in the previous section, and the ASP encoding that implements the questionnaire logic ($\Pi_q$) as described in the next paragraphs. The output is composed of the set of {\em compatible} diagnoses $C$ and the set of {\em not compatible} diagnoses $N$ according to the history of answers $H$ provided by the patient, where $H$ is a set of ASP facts. Note that, the history of answers is collected during the process. Initially, at line \ref{ln:init-U}, the algorithm retrieves the set of all possible diagnoses encoded in $\Pi_{ichd}$ and then initializes the patient's history of answers, the set of {\em compatible} answers, and the set of {\em not compatible} answers to the empty set (line \ref{ln:init-C-I}). Then, the algorithm loops until all diagnoses are determined (line \ref{ln:loop}). At each iteration, first the algorithm computes the current diagnostic status, i.e., the current set of {\em compatible} diagnoses and the current set of {\em not compatible} diagnoses, by running the \acrshort{ichd3} encoding together with the history of answers provided by the patient. To denote the fact that $\Pi_{ichd} \cup H$ has a unique {\em answer set}, we denote the solving method with $solveUnique$ (line \ref{ln:solve-unique}). At this point, the algorithm checks whether there are still diagnoses that are {\em not determined}: in the negative case it finishes, and in the positive case it computes the next question to be posed to the patient (from line \ref{ln:unfinished} to \ref{ln:add-patient-answer}). To do so, it first builds an ASP program $\Pi''$ as the union of $\Pi_{ichd}$, $\Pi_q$ and $H$. At line \ref{ln:solve}, it stores the {\em answer sets} of $\Pi''$ into a variable $\mathcal{S}$. There are $2n$ {\em answer sets}, where $n$ is the number of questions selected at the current step (further details later in this section): given a selected question $q$, there is exactly one {\em answer set} where $q$ is asked and the answer to $q$ is affirmative, and exactly one {\em answer set} where $q$ is asked and the answer to $q$ is negative. The set of selected questions is $Q$, which is computed by putting together the atoms (exactly one instance per {\em answer set}) whose predicate is $ask$(line \ref{ln:q}). From line \ref{ln:init-score} to \ref{ln:maxmin-end}, the algorithm computes, heuristically, the best question ($B_q$ in the algorithm) to pose to the patient, according to the worst-case minimization strategy discussed above. Initially, the best question $B_q$ is initialized to $\bot$ to denote the fact that it has no initial value, and the score is initialized to $-1$  (line \ref{ln:init-score}) . For each question $q$, the algorithm finds the number $d_{yes}$ of determined diagnoses of $q$ if answered affirmatively (line \ref{ln:answer-yes}) and the number $d_{no}$ of determined diagnoses of $q$ if answered negatively (line \ref{ln:answer-no}). The function $getDeterminedDiagnosesCount(q, answer, \mathcal{S})$ essentially finds the {\em answer set} in $\mathcal{S}$ where $q$ is asked and the answer to $q$ is $answer$ and returns the number of {\em determined} diagnoses in that {\em answer set}. The score of $q$ is the minimum between $d_{yes}$ and $d_{no}$. In other words, a question is scored by the number of diagnoses that are {\em determined} in the worst-case scenario. The selected question $B_q$ is the question having the maximum score (lines \ref{ln:best-start} to \ref{ln:maxmin-end}).
 Finally, the selected best question $B_q$ is posed to the patient and their answer is added to the history $H$ (line \ref{ln:add-patient-answer}): the function $patientAnswer(B_q)$ asks $B_q$ to the patient and returns their answer encoded as an ASP fact.

%The input to the logical module consists of a history of question/answer pairs which is represented in terms of ASP facts. Initially the history is empty and new facts are added to the history once the patient starts answering questions, one at a time. The output of the logical module is a set of answer sets. Each answer set contains the consequences of the answers already provided together with the answer of a new question.
%\todo{Describe the same concept in formal terms}
%DESCRIBE THE STRATEGY IN HIGH LEVEL TERMS (INPUT FACTS AND ANSWER SETS) + MIN MAX
\paragraph{ASP encoding.}
In the following, we enumerate and give an informal semantics of the predicates defined in the ASP program that implements the questionnaire described above.
Based on the notions that appear in the diagnostic criteria classified in the previous section and that will be the subject of the questions to be asked, we identified a set of topics that allow us to split the set of potential questions into groups. In the encoding that implements the questionnaire we listed such topics in the form of instances of the predicate \emph{topic(T, D)}, where, the variable \emph{T} is instantiated by the constants that identify the topics, and \emph{D} is either instantiated by the constant \emph{dependent} or \emph{independent} to indicate that a topic is related to a symptom or not. An example of a ground instance of the predicate \emph{topic} is \emph{topic(duration, dependent)}, which represents that {\em duration} is a dependent notion (e.g., the duration of headache or nausea makes sense, while duration alone does not). On the other hand, a symptom is a notion that is not dependent on another concept and we express it by the ground instance \emph{topic(symptom, independent)}.
The instances of the predicate \emph{criterionDependsOn(Id\_diag, Letter, X, Y, Topic)} express the dependence of a diagnostic criterion \emph{Letter} of a diagnosis \emph{Id\_diag}, on the elements \emph{X} and \emph{Y} that characterize the topic \emph{Topic} and that will constitute the possible subject of a question.
For example, we use  {\em criterionDependsOn (d.2.1,  ``D'', s33, ``nausea'', symptom)} to denote that the criterion ``\textsf{D}'' of the diagnosis ``\textsf{d.2.1}'' depends on the symptom ``\textsf{nausea}'', whose identifier is \emph{s33}.
Each possible subject of the question is collected in the ground instances of the predicate {\em possible} through the following rule:
\begin{equation*}
\begin{array}{l}
r_{14}: possible(X,Y,Topic)\leftarrow criterionDependsOn(Id\_diag,Letter,X,Y,Topic).
\end{array}
\end{equation*}
The predicate \emph{possible} is then used to generate the predicate \emph{ask(X, Y, Topic)} in the portion of the disjunctive program in which we evaluate asking potential questions. To the independent type topics correspond instances of the predicate {\em possible} in which \emph{X} and \emph{Y} are the characterizing elements, identified and formalized in the previous section. As an illustration, the characterizing elements of the topic \emph{exam} are its name and the relative report, so, an example of its instantiation is \emph{possible(``gene CACNA1A'', ``presence of mutation'', exam)}. In the case of independent type topic, it is necessary to represent the name of the symptom the topic refers to. Therefore, the variable \emph{X} is instantiated by the name of the symptoms and the variable \emph{Y} by the elements that characterize the topic itself. An example for the topic \emph{duration} is the atom \emph{possible(``headache'', 5, duration)} where, ``\textsf {5}'' is one of the values, belonging to the considered domain, expressed in minutes.
We define ``relevant'' an element present in a criterion relating to at least one diagnosis not yet excluded and relating to a diagnosis ``child'' (according to predicate \emph{isA}) whose diagnosis ``father'' has been confirmed.
\begin{equation*}
\begin{array}{l}
r_{15}: relevant(X, \ Y, \ Topic) \leftarrow criterionDependsOn(Id\_diag, \ Letter, \ X, \ Y, \ Topic),\\
\ \ \ \ \ \ \ \ \ \ \ \
not \ criterion(Id\_diag, \ Letter), \ not \ -diagnosis(Id\_diag), \\
\ \ \ \ \ \ \ \ \ \ \ \
isA(Id\_diag,\ Id\_sup\_diag), \ diagnosis(Id\_sup\_diag).
\end{array}
\end{equation*}
The patient's answers will be collected in the ground instances of the predicate \emph{answer(X, Y, Topic, Answer, Type)} where \emph{Answer} is either the constant \textsf{true} or \textsf{false} and the variable \emph{Type} is instantiated by the constant \textsf{real} or \textsf{simulated} to distinguish between the history of answers actually given by the patient and the answers simulated during the evaluation phase for the choice of the next question.
The logic module that implements the interactive questionnaire is structured by following the \textquotedblleft Guess and Check\textquotedblright~programming paradigm of ASP.  The \textquotedblleft Guess\textquotedblright~part of the questionnaire program is composed of a set of disjunctive rules that allow to consider the different possibilities for the choice of the next question. On the other hand, the \textquotedblleft Check\textquotedblright~part consists of a set of constraints that discard the unwanted models (e.g., models which ask more than one question simultaneously, models in which the question has already been asked or models in which the question is no more relevant). Beside disjunctive rules and constraints, ASP programs are also composed of a set of normal rules (i.e., rules with a single atom in the head) that are used to propagate knowledge in the program (e.g., propagate the consequences of an answer).
The program identifies a set of \emph{n} questions that could be asked and, in order to analyze separately the consequences of the \emph{2} possible answers from the patient, it generates \emph{2n} {\em answer sets}. In other words, for each of the \emph{n} possible questions, the program outputs an {\em answer set} that represents the outcome of the affirmative answer and an {\em answer set} that represents the outcome of the negative answer.
Overall, an {\em answer set} represents a possible world in which the system can evolve once the patient answers another question.
The following disjunctive rules compose the \textquotedblleft Guess\textquotedblright~part.
In particular,
for each topic \emph{T} considered, we choose it or we don't choose it ($r_{16}$),
for an \emph{independent} type topic $T$,
we ask, or we don't, a possible question regarding $T$ ($r_{17}$),
given a \emph{dependent} type topic \emph{T} and a possible question regarding it, after verifying the presence of the symptom the topic refers to, we ask the question or we don't ($r_{18}$),
we simulate the patient's answer to each question when it is affirmative and when it is negative ($r_{19}$), and
we exclude the evaluation, in the same set, of more than one question ($r_{20}$):
\begin{equation*}
\begin{array}{l}
r_{16}: chosenTopic(T) |  -chosenTopic (T)\leftarrow topic(T,D).\\

r_{17}: ask(X,Y,T) | -ask(X,Y,T) \leftarrow chosenTopic(T),\ possible(X,Y,T), topic(T, independent).\\

r_{18}: ask(Name,Y,T) | -ask(Name,Y,T) \leftarrow chosenTopic(T), possible(Name,Y,T),\\
\ \ \ \ \ \ \ \ \ \ \ \
topic(T,dependent), \ ichdSymptom(Id, Name),\  symptom(Id).\\

r_{19}: answer(X, Y, T, false, simulated) | answer(X, Y, T, true, simulated) \leftarrow ask(X, Y, T).\\

r_{20}: \leftarrow not\ \#count\{X,Y,T: ask(X,Y,T)\}= 1.
\end{array}
\end{equation*}
In the following, we present the main constraints that allow a resizing of the admissible questions set, in order to let them conform to the history of answers. In particular, not relevant questions cannot be asked ($r_{21}$),
a question that has been previously answered cannot be asked ($r_{22}$), and
it is not possible to consider, in a question, a numerical value for the \emph{duration} of a symptom that is not in the previously identified range ($r_{23}$):
\begin{equation*}
\begin{array}{l}
r_{21}: \ \ \leftarrow ask(X, Y, T),\ not\ relevant(X, Y, T).\\
r_{22}: \ \ \leftarrow ask(X, Y, T),\ answer(X, Y, T, \_, real).\\
r_{23}: \ \ \leftarrow ask(Name, V, duration),\ ichdSymptom(Id, Name),\ symptom(Id), \\
\ \ \ \ \ \ \ \ \ \ \ \
minDuration (Id, Y),\ V < Y.
\end{array}
\end{equation*}
The program contains also other rules similar to $r_{23}$ focusing on different numerical parameters.
By simulating the patient's answer, we evaluate the effect that such answer would have in the process of determining the diagnoses. We use the information that would be acquired through the answers to activate the corresponding predicates which, once propagated in the \acrshort{ichd3} encoding, contribute to inferring the eventual {\em compatible} and {\em not compatible} diagnoses.
The patient's affirmative answer concerning a symptom activates the predicate which models its presence.
\begin{equation*}
\begin{array}{l}
r_{24}: symptom(Id) \leftarrow answer(Id, \_ , symptom, true, \_).
\end{array}
\end{equation*}
On the other hand, the patient's negative answer concerning a symptom allows us to infer the absence (modeled with strong negation) of the predicate which models its presence.
\begin{equation*}
\begin{array}{l}
r_{25}: -symptom(Id)\leftarrow answer(Id, \_ ,  symptom, false, \_).
\end{array}
\end{equation*}
%TODO check
%While assessing the consequences of any information provided by the patient, the planning also deals with automatically managing the indirect consequences of the reported answers, as a consequence of logical considerations and deductions deriving from any hierarchical relations between the concepts characterizing the diagnostic criteria.
%Among the notions there are relations of hyponymy and mutual exclusion. The first report involves the introduction of rules which, confirmed the presence of the specific symptom by the patient, the presence of the generic symptom is inferred and, conversely, the absence of the generic symptom entails the invalidation of all the specializations related to its category.
%answer(Id2, genericSymptom, true, Topic, Type) :- answer(Id1, specificSymptom, true, Topic, Type), isHyponymSympton(specificSymptom, genericSymptom), possible(Id1, specificSymptom, symptom), possible(Id2, genericSymptom, symptom).
%answer(Id1, specificSymptom, false, Topic, Type) :- answer(Id2, genericSymptom, false, Topic, Type), isHyponymSympton(specificSymptom, genericSymptom), possible(Id1, specificSymptom, symptom), possible(Id2, genericSymptom, symptom).
Finally, we compute the number of diagnoses that would be {\em determined}. To do so, we use the aggregation construct \emph{\#count}: we count the diagnoses that would be {\em compatible} and those ones that would be {\em not compatible}.
\begin{equation*}
	\begin{array}{l}
		r_{26}: compatibleDiag(N)\leftarrow  \#count\{Id: diagnosis(Id)\} = N.\\
        	r_{27}: notCompatibleDiag(N) \leftarrow  \#count\{Id: -diagnosis(Id)\} = N.
	\end{array}
\end{equation*}
%Each {\em answer set} is, therefore, associated with the value corresponding to the sum of {\em compatible} and {\em not compatible} diagnoses. This will allow us to compare each question-answer configuration in order to choose the best question to ask at the current step.
The number of {\em determined} diagnoses is then inferred by the following rule:
\begin{equation*}
\begin{array}{l}
r_{28}: determinedDiag(D) \leftarrow compatibleDiag(N), \ notCompatibleDiag(M), \ D = N + M.
\end{array}
\end{equation*}
Such a value is then used by Algorithm~\ref{alg:questionnaire} as previously discussed.
%This concludes the rules used to identify the next question.

%\paragraph{Discussion.}

%DISCUSS  POSSIBLE OUTCOMES

\section{System implementation and testing}\label{sec:system} %BERNARDO

%VERIFICARE SE POSSIBILE INSERIRE QUALCHE ESPERIMENTO/TEST

%In this section we describe the implementation of the decision support system as a WEB-Oriented application.
The decision support system has been implemented into two Web applications: a REST Web service, and a Web graphical interface. Web applications are one of the most common types of distributed applications. In general, they are accessible without requiring any installation process and are cross-platform by design.

\begin{figure}[t!]
	\centering
	\includegraphics[width=0.8\textwidth]{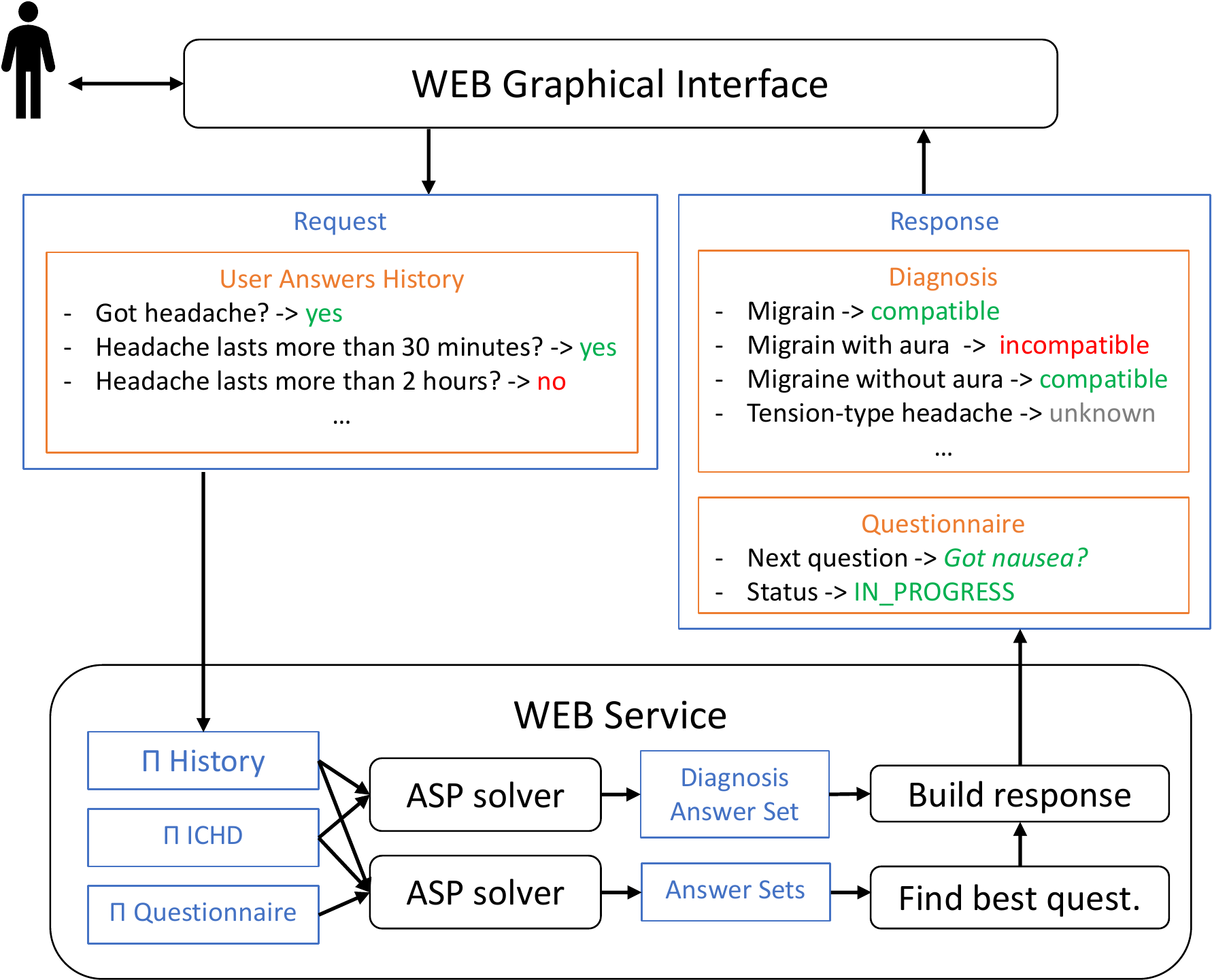}
	\caption{System architecture}\label{fig:architecture}
\end{figure}

Figure \ref{fig:architecture} presents, with an example, the architecture of Head-ASP. The user (a physician) interacts with ther Web Graphical Interface which communicates under the form of HTTP requests and responses with the Web Service.
The \textsc{head-asp} Web Service exposes to the Web the functionalities of the DSS. It provides an HTTP method that receives a user answers history and returns the current diagnosis and the next question, implementing the algorithm presented in the previous section in Algorithm \ref{alg:questionnaire}.
The Web Service is not intended to be used directly by humans, but it is instead intended to be invoked by other programs. It has been implemented in \textsc{JAVA} using {\em Spring}, which is a well-known framework for Web development. We used \textsc{DLV}~\cite{DBLP:conf/lpnmr/AlvianoCDFLPRVZ17} as the ASP solver of the application.% together with the {\em DLVWrapper} library for interacting with the ASP solver from \textsc{JAVA} code.

The implemented service follows the REST architectural style~\cite{richardson2008restful}. Web services and one of the main consequences is that it is stateless, meaning that it has no concept of state or session, and thus it can scale well horizontally~\cite{michael2007scale}: more instances can be deployed simultaneously and which instance is handling a request is not important. The Web service also exposes a documentation that shows its communication protocol, i.e., how to invoke it and how to interpret its output. In simple terms, the Web service exposes a method that accepts a patient history of answers and returns the current diagnosis and the next question. The response contains special tokens in the cases where the questionnaire is completed (i.e., every diagnosis is either {\em compatible} or {\em not compatible}) or the questionnaire can not be continued (i.e., there is no relevant question to ask).

The \textsc{head-asp} Web graphical interface demonstrates the functionalities of the DSS and, contrary to the Web service, it is intended to be used by humans. Currently, it is used as the primary interface of \textsc{head-asp} until the future integration of the DSS in the {\em Alcmeone} project platform. To develop the Web interface we used {\em Angular 8}, which is a popular {\em JavaScript} framework for the development of Web graphical interfaces.
The Web interface and the Web service (and its documentation) are available at \url{https://head-asp.github.io/ichd-dss/}.

\begin{figure}
\subfigure[Reaching the complete diagnostic picture.]
{\includegraphics[width=6.5cm]{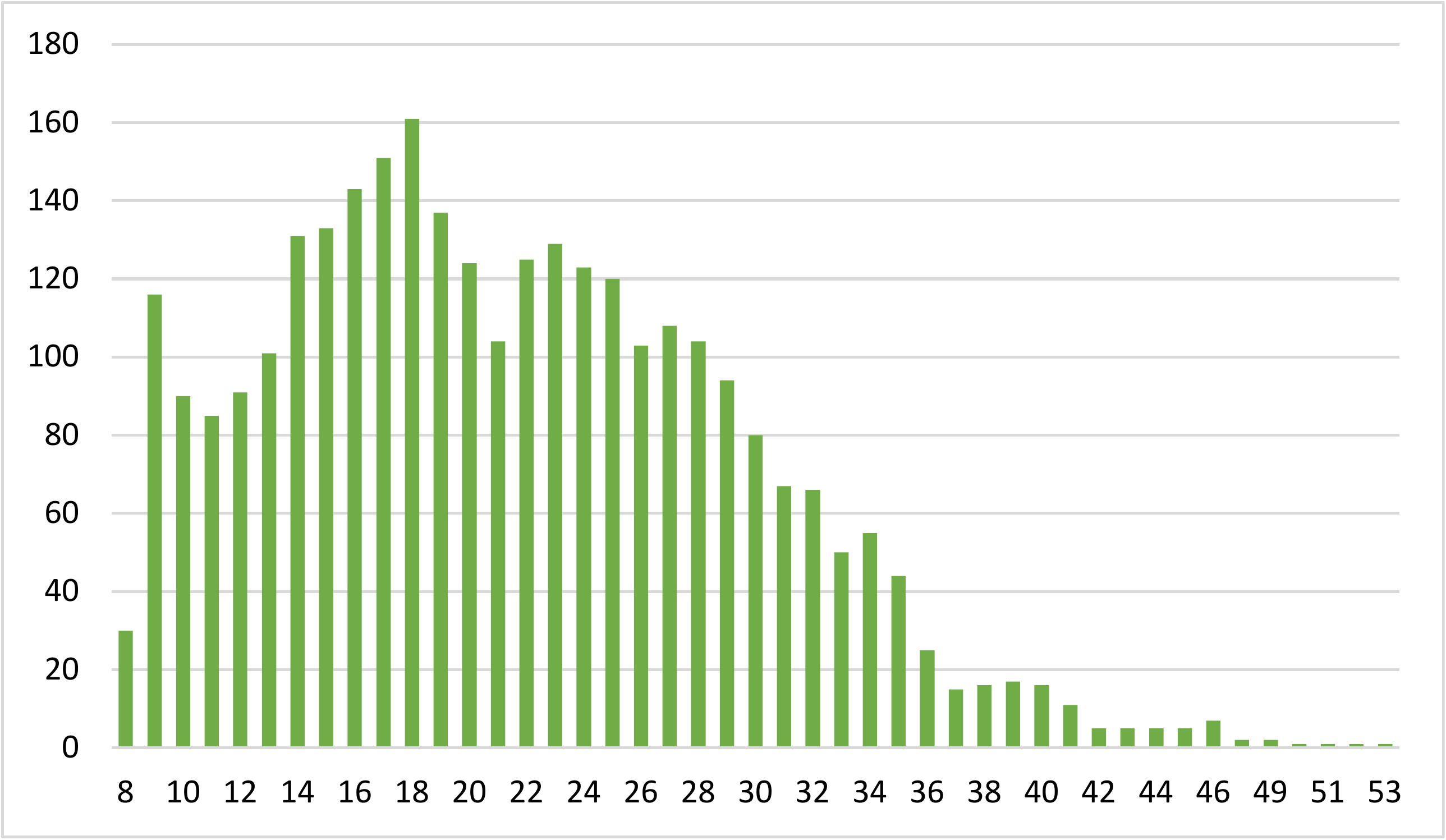}}
\hspace{0.5em}
\subfigure[Reaching the first compatible diagnosis.]
%Questionnaire length for first compatible diag.
{\includegraphics[width=6.5cm]{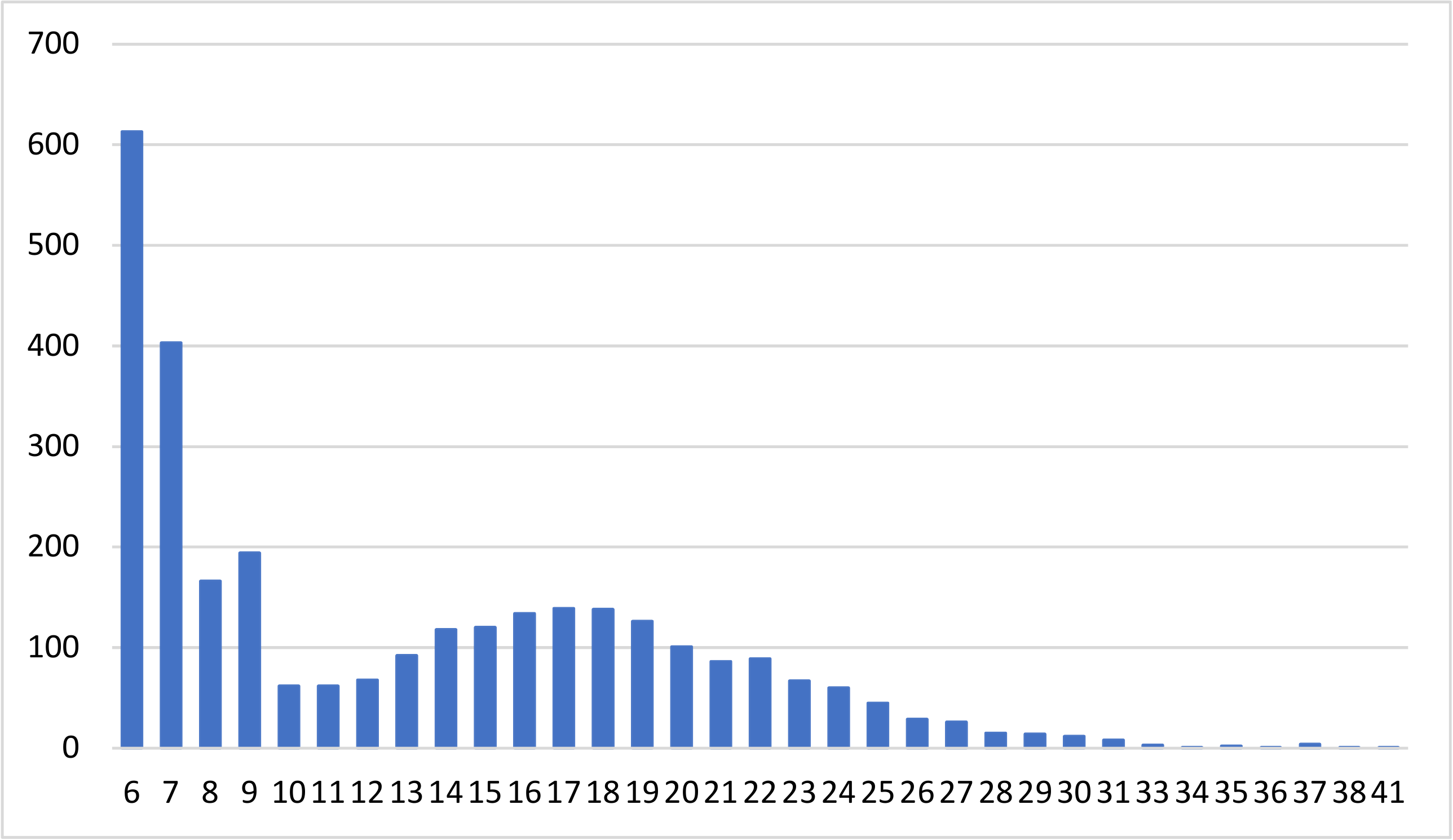}}
\caption{Distribution of the number of questionnaires by length under random answers.}
%	Results of 3000 randomized tests with at least one compatible diagnosis}
\label{fig:experiments}
\end{figure}

%Distribution of people involved in the survey by age

To minimize the number of defects in the implementation and investigate the performance of the approach, we implemented a testing framework where questions are answered randomly. In this section, we report the results obtained by the current release of \textsc{head-asp} on approximately 7400 questionnaires. Among these, we focused on those --3000 in total-- that led to at least one headache disorder, which are representative of real cases of headache. Figure~\ref{fig:experiments} shows two histograms summarizing the results of our tests.
In both histograms, each bar on top of a value $x$ represents how many questionnaires had a length $x$.	
In $(a)$, we plot the distribution of the number of questionnaires by overall length (i.e., the length necessary to reach a complete diagnostic picture); in $(b)$, we consider the distribution of the number of questionnaires by the length at which the first compatible diagnosis has been reached. The average length is of 21.44 questions in the first case and of 12.99 in the second one.
It is worth noting that these results are in line with the neurologists' experience and expectations. Finally, considering the fact that there are more than 150 candidate questions encoded in the system,
the results of our experiment reveal how \textsc{head-asp} is able to effectively discard unnecessary questions when needed, pruning the search space and producing short questionnaires.

\section{Discussion and future work}
In this paper, we have presented {\textsc{head-asp}}: a novel decision support system for the diagnosis of headache disorders ---one of the most common and disabling conditions of the nervous system throughout the world.
The system is currently tested by a group of neurologists that are profitably using it within {\em Alcmeone} ---a research project whose objective is providing an innovative organizational and management model, and an advanced technological platform of services for supporting the integrated clinical management of headache patients.

Although design and implementation of the DSS was quite challenging, we found very natural to use logic programming both to encode the ICHD classification and to model the heuristics for determining the next question. In particular, the advantages can be summarized as follows:
	$(i)$ simple and complex diagnoses can be encoded in a natural and precise way;
	$(ii)$ ICHD updates can be easily and locally transferred to the encoding;
	$(iii)$ medical knowledge can be represented and integrated in a declarative way; and
	$(iv)$ identifying candidate questions and simulating their effects are tasks that can be carried out in the same framework.
	Moreover, concerning the adoption of ASP w.r.t. other languages such as Prolog, we feel that a bottom-up paradigm here is more appropriate since we start from some domain knowledge, we add patient's symptoms, we derive compatible and not compatible criteria, and we identify compatible and not compatible diagnoses. Moreover, we do not have a specific query to evaluate but each time we have to update our diagnostic picture marking each diagnosis as compatible, not compatible or not-yet-determined.

Concerning our future plans, the ultimate objective is
to promote the system inside the Italian Society for the Study of Headache (SISC) so that it can become a valid support to clinicians and specialists. To this end, we are still refining and improving it according to the feedback we are currently receiving. In particular, the next steps include:
$(i)$ completing the ICHD-3 encoding;
$(ii)$ analyzing the next-question problem from a theoretical perspective;
$(iii)$ further reducing the average number of questions needed to reach a complete diagnosis.
Least but not last, we would like to generalize our methodology to be easily applicable in similar contexts.

%\appendix

%\section{\add{Further details of the logical representation}}

%\add{TODO}

%\section{\add{Formal architecture of the questionnaire}}

%\add{TODO}

\section*{Acknowledgments}
The authors want to thank Rosario Iannacchero (Neurologist and headache specialist at the Headache Regional Center, Division of Neuroloy, “Pugliese-Ciaccio” Hospital, Catanzaro, Italy), Domenico Conforti (Full professor at the Department of Mechanical, Energy, Management Engineering, University of Calabria, Rende, Italy) and Giovanni Laboccetta (Chief technology officer at DLVSystem Srl, Rende, Italy) for useful discussions and suggestions during the development of \textsc{head-asp} and for their precious feedback.

% for suggestions and scientific advice during the various project development stages
\bibliographystyle{style/acmtrans}
\bibliography{references}

\end{document}